\documentclass{article}

\usepackage[preprint]{neurips_2021}
\usepackage{amsmath}
\usepackage{graphicx}
\usepackage{enumitem}

\graphicspath{ {./img/} }

\usepackage[utf8]{inputenc} % allow utf-8 input
\usepackage[T1]{fontenc}    % use 8-bit T1 fonts
\usepackage{hyperref}       % hyperlinks
\usepackage{url}            % simple URL typesetting
\usepackage{booktabs}       % professional-quality tables
\usepackage{amsfonts}       % blackboard math symbols
\usepackage{nicefrac}       % compact symbols for 1/2, etc.
\usepackage{microtype}      % microtypography
\usepackage{xcolor}         % colors

\newcommand{\lauro}[1]{\textcolor{blue}{\textbf{lauro}: #1}}
\newcommand{\edouard}[1]{\textcolor{green}{\textbf{edouard}: #1}}

\renewcommand{\lauro}[1]{}
\renewcommand{\edouard}[1]{}

\title{A technical note on bilinear layers for interpretability}

\author{%
Lee Sharkey\\
\texttt{\{lee@conjecture.dev, leedsharkey@gmail.com\}}\\
Conjecture
}

\begin{document}

\maketitle

\begin{abstract}
The ability of neural networks to represent more features than neurons makes interpreting them challenging. This phenomenon, known as superposition \citep{olah2020zoom, toy_models_superposition}, has spurred efforts to find architectures that are more interpretable than standard multilayer perceptrons (MLPs) with elementwise activation functions. In this note, I examine bilinear layers \citep{shazeer_gated}, which are a type of MLP layer that are mathematically much easier to analyze while simultaneously performing better than standard MLPs. Although they are nonlinear functions of their input, I demonstrate that bilinear layers can be expressed using only linear operations and third order tensors. We can integrate this expression for bilinear layers into a mathematical framework for transformer circuits \citep{math_framework}, which was previously limited to attention-only transformers. These results suggest that bilinear layers are easier to analyze mathematically than current architectures and thus may lend themselves to deeper safety insights by allowing us to talk more formally about circuits in neural networks. Additionally, bilinear layers may offer an alternative path for mechanistic interpretability through understanding the \textit{mechanisms of feature construction} instead of enumerating a (potentially exponentially) large number of features in large models.
\end{abstract}

\section{Introduction} 
\label{introduction}
Neural networks can learn to compute interesting and complicated functions. To a first approximation, these functions appear to be structured such that particular computational roles or representations are assigned to particular directions in neural activation space \citep{olah2020zoom}. We call these representations \textit{features}. Somewhat surprisingly, neural networks are believed to be able to represent more features than they have neurons \citep{toy_models_superposition, gurnee2023}. This phenomenon is known as \textit{superposition}, since they assign features to non-orthogonal directions which `overlap' in high-dimensional space. 

We are particularly interested in mechanistically understanding large language models that use the transformer architecture \citep{vaswani2017}. This architecture mostly consists of a series of alternating attention layers (which let activations at different points in a sequence interact with each other) and MLP layers (which, at each point in the sequence, construct useful output features that are nonlinear transformations of the input features). About two thirds of the parameters in these models are in the MLP layers, which are thought to make prodigious use of superposition \citep{elhage2022solu, gurnee2023}. 

Nonlinear elementwise activation functions (such as ReLU \citep{nair_hinton_relu_2010} or GeLU \citep{hendrycks2020gaussian}) in MLP layers can remove small amounts of interference between non-orthogonal features \citep{toy_models_superposition}, thus making it possible for layers to represent features in superposition without increasing the loss. Unfortunately, while the activation function is very useful for the performance of neural networks, it makes it quite difficult to analyze MLPs mathematically because the powerful tools of linear algebra can no longer be readily applied. 

However, it turns out that another kind of MLP layer, the bilinear layer \citep{shazeer_gated, dauphin2016, mnih2007}, is much easier to analyze than MLPs with elementwise activation functions. Even though bilinear layers are nonlinear functions of the input vector, \textbf{bilinear layers can be described using only linear operations and third order tensors}! This nice property lets us \textbf{extend `A Mathematical Framework for Transformer Circuits' \citep{math_framework} to transformers with MLP layers as well as attention}, not just attention-only transformers. We hope that this simple change will give us a firmer analytical footing to understand large models on a deep, mechanistic level. This might eventually let us make deeper claims about their safety, since it could permit us to describe classes of circuits as mathematical objects with certain properties (as induction heads were in \cite{math_framework}) and to analyze learning dynamics and predict the emergence of particular kinds of circuits.

It has been hypothesized (though not yet observed) that neural networks might represent a number of features that is exponential in the number of neurons in a layer \citep{toy_models_superposition}. If this is true, it would not bode well for our ability to mechanistically understand large neural networks, which in a sense relies on our being able to enumerate all their features. However, as discussed in the last section of this work, bilinear layers may offer \textbf{a potential alternative path to `enumerative safety'} \citep{toy_models_superposition}. Instead of attempting to understand each of a large number of features, with bilinear networks we may be able to understand a smaller number of primitive features that bilinear layers use to `construct' their (potentially exponential) larger number of features. Thus, in the same way that we might be able to understand an exponentially large number of executed programs by understanding their code, we might be able to understand an exponentially large number of features by understanding the process by which features with certain properties are constructed. Here, we make some preliminary steps toward understanding the mechanisms of feature construction in bilinear layers; we show that in bilinear layers, \textbf{output features are constructed through sums of pairwise interactions between input features}, whereas, in standard MLPs, output features are constructed using all-to-all interactions between input features that appear not to be decomposable.

\section{Bilinear layers} \label{bilinear_intro}

\subsection{Introducing bilinear layers}

A standard MLP layer consist of an input vector $x$, a weight matrix $W$ and an elementwise nonlinear activation function, $\sigma$ such as the ReLU function (and an optional bias term which is omitted for notational simplicity). The input vector is linearly transformed by the weight matrix to yield the pre-activation vector $W x$, to which the activation function is applied elementwise:
$$MLP_{ReLU}(x) = \sigma(W x)$$

Bilinear layers are slightly different. They take the form 
   $$MLP_{Bilinear}(x) = (W_1 x) \odot (W_2 x),$$
where $\odot$ denotes elementwise multiplication. They have two weight matrices, which each separately transform the input vector. They were introduced in different forms by \cite{dauphin2016} and \cite{mnih2007}. They were later studied by \cite{shazeer_gated}, who showed that bilinear layers, when used as the MLP layer in transformer language models, are surprisingly competitive\footnote{At least for the model size they explored, which was approximately 120M parameters, a similar size to GPT2-small \citep{radford2019}. To my knowledge, it remains to be determined whether bilinear layers continue to perform competitively at larger scales.}: They are at least as performant per parameter than standard MLPs with ReLU or GELU activation functions and only slightly less performant than state-of-the-art SwiGLU layers\footnote{A SwiGLU layer is equivalent to a bilinear layer but where an elementwise Swish activation function \citep{ramachandran2017searching} is applied to $W_1 x$.}.

\subsection{Describing bilinear layers using only linear operations and third order tensors}

The lack of an elementwise activation function in bilinear layers makes them mathematically very simple. In fact, despite being nonlinear functions of $x$, they can be expressed using only linear operations and third order tensors. 

First, we'll define the \textit{tensor inner product} (See appendix \ref{sec:inner_prod_examples} for some examples of tensor inner products which may help build intuitions). Unlike the inner product between vectors, the tensor inner product needs to define the axes along which the inner product is taken. The tensor inner product is thus defined as
$$\boldsymbol{U}^{(n)} \boldsymbol{\cdot}_{jk} \boldsymbol{V}^{(m)} = \boldsymbol{T}^{(n+m-2)}$$ where 

\begin{multline}
\boldsymbol{T}_{\gamma_1 \cdots \gamma_{j-1} \gamma_{j+1} \cdots \gamma_{n}  \gamma'_{1} \cdots \gamma'_{k-1} \gamma'_{k+1} \cdots \gamma'_{m} } = \sum_\beta \boldsymbol{U}_{\gamma_1 \cdots \gamma_{j-1} \beta \gamma_{j+1} \cdots \gamma_n}  \boldsymbol{V}_{\gamma'_{1} \cdots \gamma'_{k-1} \beta \gamma'_{k+1} \cdots \gamma'_{m}}
\end{multline}

For the tensor inner product between $n^{\text{th}}$ order tensor $\boldsymbol{U}$ and $m^{\text{th}}$ order $\boldsymbol{V}$ to be defined, the dimension of axis $j$ of tensor $\boldsymbol{U}$ must be the same dimension as axis $k$ of tensor $\boldsymbol{V}$.

Now we show how bilinear layers can be expressed using linear operations and third order tensors. Suppose we want to find the third order tensor $B$ such that 
$$(W_1 x) \odot (W_2 x) = x \boldsymbol{\cdot}_{1 2} B \boldsymbol{\cdot}_{2 1} x ,$$ if it exists. 
We'll first identify the terms in the vector on the right hand side, 
\begin{equation} \label{bilin_rexpr}
\begin{split}
((W_1 x) \odot (W_2 x) )_i & = (\sum_j W_{1 (i j)} x_j)(\sum_{k} W_{2 (i k)} x_{k}) \\
& = \sum_j \sum_{k} W_{1 (i j)} x_j W_{2 (i k)} x_{k}
\end{split}
\end{equation}

Now let's express the terms of the third order tensor $B$ using tensor inner products. We have,
\begin{equation}\label{B_i_term}
\begin{split}
(x \boldsymbol{\cdot}_{1 2} B \boldsymbol{\cdot}_{2 1} x)_i = \sum_j x_j \sum_{k} x_{k} B_{i j k} \\  = \sum_k x_k \sum_{j} x_{j} B_{i j k} \\ = \sum_j  \sum_{k} x_j x_{k} B_{i j k}.
\end{split}
\end{equation}

Note that it doesn't matter whether we take the tensor inner product between $B$ and $x$ on the $2$nd or $3$rd axis first, which is why $x \boldsymbol{\cdot}_{1 2} B \boldsymbol{\cdot}_{2 1} x$ is associative, i.e. $(x \boldsymbol{\cdot}_{1 2} B ) \boldsymbol{\cdot}_{2 1} x = x \boldsymbol{\cdot}_{1 2} ( B \boldsymbol{\cdot}_{2 1} x)$. We'll use this property when extending a Mathematical Framework for Transformer Circuits \citep{math_framework} (Section \ref{extending_math_framework}).

Comparing the terms from equations \ref{bilin_rexpr} and \ref{B_i_term}, we can see they are equal if
$B_{i j k} = W_{1 (i j)} W_{2 (i k)}.$ Thus, we can construct the tensor $B$ using the bilinear layer weights $W_1, W_2 \in \mathbb{R}^{m \times n}$ and a third order tensor $Z$ such that $Z_{ijk} = 1$ where $i=j=k$ and $0$ otherwise, because $B = W_1 \boldsymbol{\cdot}_{1 2} Z \boldsymbol{\cdot}_{2 1} W_2 $. One helpful way to think about the $m \times n \times n$ tensor $B$ is that the column vector $B_{: j k}$ consists of the elementwise multiplication of the $j^{\text{th}}$ column of $W_1$ with the $k^{\text{th}}$ column of $W_2$.

\subsection{Extending a Mathematical Framework for Transformer Circuits}\label{extending_math_framework}

When \cite{math_framework} analyzed the equations for 1- and 2-layer attention-only transformers, it offered interesting insights on the structure of these models. It helped to reveal QK- and OV-circuits, induction heads, and virtual attention heads, which formed the basis of much interesting follow-up work in interpretability \citep{olsson2022incontext, wang2022interpretability}.

However, one of the biggest shortcomings of \cite{math_framework} was that the transformers they analyzed had no MLP layers. MLPs comprise around two thirds of all parameters in standard transformer language models and are thought to be be necessary for a great deal of interesting behaviour \citep{geva2021transformer}. The reason MLPs were excluded was that they could not be linearised, which made their analysis intractable. But, as we've seen, it is possible to describe bilinear layers using only linear operations. This means we can write linearized expressions for transformers with both attention and MLP layers! It's important to stress that the MLPs we achieve this with are close to state of the art \citep{shazeer_gated}. This opens up the possibility that we may be able to formally analyze some very capable language models. 
In this section, we'll identify the expression for a one-layer transformer with attention and (bilinear) MLPs. The expressions for two- and N-layer transformers are left as lengthy exercises for the reader. 

We'll update our notation in order to be consistent with \cite{math_framework}, with which we expect readers to be familiar. The inputs to the language model is a sequence of tokens $t$ of length $n_{\text{context}}$. These are embedded by the ${d_{\text{model}} \times n_{vocab}}$ embedding matrix $W_E$. The token embeddings $x_0 = W_E t$ (which have shape ${n_{\text{context}} \times d_{\text{model}}}$ ) become the residual stream, which is passed through multiple residual blocks, each consisting of a multihead attention layer and an MLP layer, and each added back into the residual stream. Finally, the residual stream is unembedded by the unembedding matrix $W_U$ to make the token logits.

In \cite{math_framework}, they assumed MLPs that had an elementwise GeLU activation function, which are very difficult to analyze. Here, we'll instead use bilinear layers. Define the bilinear MLP layer as
\begin{equation}
\begin{split}
F(x) &= W_O^{m}(x \boldsymbol{\cdot}_{1 2} W_{I_1}^{m} \boldsymbol{\cdot}_{1 2} Z \boldsymbol{\cdot}_{2 1} W_{I_2}^{m} \boldsymbol{\cdot}_{2 1} x)
\end{split}
\end{equation}
where 
$W_{O}^{m}$ is the $d_{\text{model}} \times d_{\text{mlp}}$ output weight matrix for the MLP layer and
$W_{I_1}^{m}, W_{I_2}^{m}$ are the two $d_{\text{mlp}} \times d_{\text{model}}$ input weight matrices for the bilinear layer.

Using the path expansion trick described by \cite{math_framework}, the input to the MLP in a one layer transformer can be described as
\begin{equation}
\begin{split}
x_1 & = (Id + \sum_{h \in H} A^h \otimes W_{OV}^h) \cdot W_E t \\
&  = (W_E + \sum_{h \in H} A^h \otimes W_{OV}^h W_E) t
\end{split}
\end{equation}

where 
$W_{OV}^h = W_{O}^h W_{V}^h$ and
$A^h = \text{softmax*}( t^T \cdot W_E^T W_{QK} W_E \cdot t)$
in which $\text{softmax*}$ is the softmax function with autoregressive masking and $W_{QK} = W_Q^{h \top} W^h_K$.
Putting our definition of $x_1$ into our definition of $F(\cdot)$ we get 
\begin{multline}
F(x_1) = W_O^{m}(((W_E + \sum_{h \in H} A^h \otimes W_{OV}^h W_E) t) \boldsymbol{\cdot}_{1 2}  W_{I_1}^{m} \boldsymbol{\cdot}_{1 2} Z \boldsymbol{\cdot}_{2 1} W_{I_2}^{m} \boldsymbol{\cdot}_{2 1} \\ ((W_E + \sum_{h \in H} A^h \otimes W_{OV}^h W_E) t))
\end{multline}
Note that for arbitrary matrices $M$, $M'$, it's true that $M \boldsymbol{\cdot}_{1 2} M' = M^\top M'^\top$. So we transpose the left hand bracket and $W_{I_1}^{m}$ and move the weight matrix into the brackets:

\begin{multline}
   = W_O^{m}((t^\top (W_E^\top W_{I_1}^{m \top} + \sum_{h \in H} A^h \otimes W_E^\top W_{OV}^{h \top} W_{I_1}^{m \top})) 
   \boldsymbol{\cdot}_{1 2} Z \boldsymbol{\cdot}_{2 1} W_{I_2}^{m} \boldsymbol{\cdot}_{2 1} \\ ((W_E + \sum_{h \in H} A^h \otimes W_{OV}^h W_E) t))
\end{multline}

And next,  noting that $M \boldsymbol{\cdot}_{2 1} M' = M M'$, we move $W_{I_2}^{m}$ into the right hand brackets:

\begin{multline}
   = W_O^{m}((t^\top (W_E^\top W_{I_1}^{m \top} + \sum_{h \in H} A^h \otimes W_E^\top W_{OV}^{h \top} W_{I_1}^{m \top}))   \boldsymbol{\cdot}_{1 2}
   Z \boldsymbol{\cdot}_{2 1}  \\
   ((W_{I_2}^{m} W_E + \sum_{h \in H} A^h \otimes W_{I_2}^{m} W_{OV}^h W_E) t))
\end{multline}

Next, we move the $Z$ tensor into the left hand brackets

\begin{multline}
   = W_O^{m}((t^\top (W_E^\top W_{I_1}^{m \top} \boldsymbol{\cdot}_{1 2} Z + 
   \sum_{h \in H} A^h \otimes W_E^\top W_{OV}^{h \top} W_{I_1}^{m \top} \boldsymbol{\cdot}_{1 2} Z))  \boldsymbol{\cdot}_{2 1} \\
   ((W_{I_2}^{m} W_E +
   \sum_{h \in H} A^h \otimes W_{I_2}^{m} W_{OV}^h W_E) t))
\end{multline}

And combining both the left hand and right hand brackets, we get the expression for a bilinear feedforward layer

\begin{equation}
\begin{split}
= & W_O^{m}(t^\top ( \\
   & \textcolor{red}{W_E^\top W_{I_1}^{m \top} \boldsymbol{\cdot}_{1 2} Z \boldsymbol{\cdot}_{ 2 1 } W_{I_2}^{m} W_E} + \\
   & \textcolor{blue}{\sum_{h \in H} A^h \otimes (W_E^\top W_{OV}^{h \top} W_{I_1}^{m \top} \boldsymbol{\cdot}_{1 2} Z  \boldsymbol{\cdot}_{ 2 1 } W_{I_2}^{m} W_E)}  +\\
   & \textcolor{magenta}{\sum_{h \in H} A^h \otimes (W_E^\top W_{I_1}^{m \top} \boldsymbol{\cdot}_{1 2} Z  \boldsymbol{\cdot}_{ 2 1 } W_{I_2}^{m} W_{OV}^{h \top} W_E)}  +\\
   & \textcolor{purple}{\sum_{h \in H} \sum_{h' \in H} A^h A^{h'} \otimes (W_E^\top W_{OV}^{h \top} W_{I_1}^{m \top} \boldsymbol{\cdot}_{1 2} Z  \boldsymbol{\cdot}_{ 2 1 } W_{I_2}^{m} W_{OV}^{h' \top} W_E)} \\
   &)t)\\
\end{split}
\end{equation}

We can analyze each of the terms in this equation. 
The \textcolor{red}{first summand} expresses a direct path from the token embedding matrix straight to the MLP without passing through any attention heads. 
The \textcolor{blue}{second summand} expresses the components of the token embeddings that pass through the attention head and then pass into only the first MLP input matrix.
The \textcolor{magenta}{third summand} is similar, but the embeddings pass through the attention heads and into the second MLP input matrix. 
The \textcolor{purple}{last summand} corresponds to token embeddings that pass through the attention heads and then into both the first and second MLP input matrices. 

With this expression for the MLP layer, we can now express the path expansion for the full one layer transformer, which is simply the above expression for $F(x)$ added to the \textcolor{orange}{token embedding-unembedding pathway (the `direct pathway')} and the \textcolor{violet}{pathways through the attention heads}:

\begin{equation}
\begin{split}
   T(t) = & \textcolor{orange}{(Id \otimes W_U W_E)t} + \\
   & \textcolor{violet}{\sum_{h \in H} (A^h \otimes W_U W_{OV}^h W_E) t} +\\
   & W_O^{m}(t^\top ( \\
   & W_E^\top W_{I_1}^{m \top} \boldsymbol{\cdot}_{1 2} Z \boldsymbol{\cdot}_{ 2 1 } W_{I_2}^{m} W_E + \\
   & \sum_{h \in H} A^h \otimes (W_E^\top W_{OV}^{h \top} W_{I_1}^{m \top} \boldsymbol{\cdot}_{1 2} Z  \boldsymbol{\cdot}_{ 2 1 } W_{I_2}^{m} W_E)  +\\
   & \sum_{h \in H} A^h \otimes (W_E^\top W_{I_1}^{m \top} \boldsymbol{\cdot}_{1 2} Z  \boldsymbol{\cdot}_{ 2 1 } W_{I_2}^{m} W_{OV}^{h \top} W_E)  +\\
   & \sum_{h \in H} \sum_{h' \in H} A^h A^{h'} \otimes \\
   & \qquad (W_E^\top W_{OV}^{h \top} W_{I_1}^{m \top} \boldsymbol{\cdot}_{1 2} Z  \boldsymbol{\cdot}_{ 2 1 } W_{I_2}^{m} W_{OV}^{h' \top} W_E) \\
   &)t)\\
\end{split}
\end{equation}

\section{Understanding feature construction in bilinear layers} \label{feat_construct}

One of the problems we may face when trying to mechanistically understand neural networks is that they may be able to represent an exponential number of features. If this hypothesis resolves true, then enumerating all the features in large networks may become computationally intractable. One analogy that gives us hope is discussed by \cite{olah2022}: Even though the input space to a particular computer program might be exponentially large, we can still say that we understand that exponentially large space of executed programs if we understand its code. In the same way, if we can understand the process by which features with certain properties are constructed from simpler primitives, we may be able to overcome the issue of having to understand an exponential number of features. In this section, which is more speculative than earlier sections, I outline why this hopeful vision seems very hard to realise in standard MLPs, but seems quite possible in bilinear layers.

\subsection{Feature construction in standard MLPs is non-decomposable} \label{MLP_monolith}
Suppose we have a standard MLP layer $MLP_{ReLU}(x) = \sigma(W x)$
with a ReLU activation $\sigma$ (where the bias term is omitted). Also suppose that the input vector $x\in X$ consists of sparse linear combinations of input features $x =  D^{I \top} a^{I}$, where $D^{I}$ is a dictionary of input features represented as a $n_{\text{features}} \times d_{\text{input}}$ matrix and $a^{I}\in A^I$ is a sparse vector of coefficients (with values in $[0, \infty)$ of size $n_{\text{features}}$) such that the dataset $X$ can be reconstructed from the features and their coefficients, $X= D^{I \top} A^{I }$. Similarly suppose there is a dictionary of output features for this layer $D^{O}$ and that sparse linear combinations of those output features describe the activations observed in a large representative sample from $p_x(MLP_{ReLU}(x))$, i.e. 
\begin{equation}
    MLP_{ReLU}(x) = \sigma(W x) = \sigma(W (D^{I \top} a^{I})) = D^{O \top} a^{O} 
\end{equation}
Therefore $D^{I}$ and $D^{O}$ are overcomplete bases\footnote{In linear algebra, a basis of a vector space is a set of vectors from which every vector in that space can be expressed as a linear combination. An \textit{overcomplete} basis is a basis where at least one element of the basis set can be removed yet the set remains a basis.} for the input space $X$ and output space $MLP_{ReLU}(X)$ respectively.  
%This somehow happens without interference between features even they may be non-orthogonal with each other\footnote{The assumption that interference is low is based on the idea that, if there were interference, then loss would be higher. Trained networks, therefore, learn to structure their features such that interference is minimized \cite{toy_models_superposition}.}. We want to tell decomposed (or `mechanistic') stories for how neural networks achieve this.

One way to view the process of feature construction is to say that output features $D^{O}$ are all implicitly represented in superposition in the weight matrix $W$ and that the nonlinearity, when applied elementwise to the preactivation vector $W x $, modifies a set of \textit{default output features} in order to select particular output features. One candidate for the default output features are the left singular vectors of $W$, i.e. the columns of a matrix $U$  (We'll discuss other candidates in the next section). We can thus introduce a \textit{modifier vector} $m(x)$ that is a function of $x$ such that 
$$ MLP_{ReLU}(x) = m(x) \odot W x = (m(x) \odot U ) \Sigma V^{\top} x = D^{O \top} a^{O}.$$ 

Therefore we can view linear combinations of the output features (namely $D^{O \top} a^{O}$) as consisting of linear combinations of modified default output features (namely $(m(x) \odot U ) \Sigma V^{\top} x$).

With a ReLU activation function, $m(x)$ is binary vector of ones and zero: $m(x)_i = 1$ where $\sigma(W x)_i > 0$ and $m(x)_i = 0$ otherwise. In general, for vanilla MLPs with any elementwise activation function $\sigma$:
\begin{equation}\label{modif_vanilla}
m(x)_i = \frac{\sigma(Wx)_i}{(Wx)_i} \quad \footnote{Note that $m(x)_i $ is discontinuous at $(Wx)_i = 0$.}
\end{equation}

It is the modifier vector that `selects' from the features represented in superposition in $W$, or, equivalently, `contructs' them by modifying the default output features. If we could understand how $m(x)$ is computed in terms of input features $D^{I}$, then we could begin to understand why particular output features $D^{O}$ are constructed not others. Unfortunately, in vanilla MLPs, the only way to calculate the value of $m(x)$ in general is Equation \ref{modif_vanilla}. In other words, to get the value of the modifier vector, we first have to pass the input through the network to observe what the post-activations (the numerator) and pre-activations are (the denominator) to get $m(x)$. But this is circular: We would have to already understand the nonlinear computation in the numerator in order to understand how output features are constructed. This framing doesn't simplify anything at all! Feature construction in standard MLPs can thus be considered `non-decomposable'.

\subsection{Feature construction in bilinear layers} \label{MLP_monolith}

In mechanistic interpretability, one of the major assumptions that we need to make is that we can interpret linear transformations of almost arbitrary dimensionality. They may still be large objects, but linear transformations are as simple as transformations get. For large linear transformations with non-sparse coefficients, we may have to spend more time studying them or prioritize analysis of the largest coefficients. But overall we assume that we can understand them to a satisfying extent. If we can't, then the whole business of mechanistic intepretability would be doomed even for large linear regressions, never mind deep neural networks.

Granting this assumption, if we could describe the modifier vector $m(x)$ in the previous section as a linear function of input features (instead of a nonlinear one), then we could begin to understand how a layer constructs output features. Fortunately, in bilinear layers the modifier vector is a linear function of the input! 
   $$MLP_{Bilinear}(x) = m(x) \odot (W_2 x) \qquad \text{where} \qquad m(x)=W_1 x,$$
We'll say that the modifier vector
modifies the default output features represented in $W_2$ to construct output features.

We still need to define what the default output feature directions and the modifier feature directions are concretely. Ultimately this choice will always be somewhat arbitrary because linear transformations do not imply any particular privileged basis.
As before, perhaps the most obvious candidates for the default output feature directions are the left singular vectors of $W_2$. But the largest directions in the output activations may not necessarily have a strong relationship with the weights because the output directions depend on both the weights and the input directions. 
Therefore, we may be able to do better than the left singular vectors of $W_2$ by incorporating the data distribution into the choice of bases. One way might use the right singular vectors $MLP_{Bilinear}(X)$ or $W_2 X$. Another -- perhaps better -- way is to identify default output features that are maximally statistically independent. This may be better because statistically independent directions tend to be activated somewhat sparsely and therefore might be better default output features than singular vectors, since fewer will be significantly `activated' at any one time. We could achieve this by performing linear ICA \cite{ica} on the preactivations $W_2 X$. This would yield a matrix $U^{(2)}$, which is the set of vectors that are maximally statistically independent directions of the output dataset while still being a basis of it. We can then use multiple linear regression to find the corresponding matrix $V^{(2) \top}$ such that $W_2 = U^{(2)} V^{(2)\top}$. Slightly abusing terminology, we'll call $U^{(2)}$ and $V^{(2)\top}$ the left and right independent components of $W_2$ respectively. We can define the modifier features using the same procedure, identifying the left and right independent components of $W_1 = U^{(1)} V^{(1)\top}$.

Armed with such features, \textbf{we may be able to describe feature construction in bilinear networks in terms of interactions between two relatively small, relatively sparse sets of vectors (the default output features and the modifier features)}. We hope we can use this approach to tell mechanistic stories for how features with certain properties are constructed by the network. We might be able to do this by understanding the functional properties of the default output features and how modifier features tend to modify them. Optimistically, mechanistic stories like these may let us understand an exponentially large space of features. Whether or not such an approach will work is ultimately an empirical question, which we leave for future work. In the next section, we explore the mathematical simplicity of feature construction in bilinear layers, which gives us some reason to suspect that feature construction may be understandable. \footnote{
We can make further modifications to the modifier features and default output features that assist either the intuitiveness or interpretability of bilinear networks. I'll note them here but won't explore them further in this work. 

\textbf{Improving intuitiveness}:  If, during training, we constrain $W_1 x$ to be low $L_2$ norm and add the one vector as bias, the modifier vector would always be close to the one vector. In other words: $m(x) = W_1 x + \textbf{1}$ where $||W_1 x|| \approx 0$. This would mean that modifier features simply cause slight modifications of default output features. This addition would also help us make a analysis prioritization decisions later (see section \ref{how_to_study_interactions}), but fundamentally the modification isn't necessary. This addition also opens up an experimental avenue (which we won't explore here): By imposing more or less regularization on the norm, it allows us to control the amount of superposition a network is able to do. This would be an interesting experimental lever to pull, since it would allow us to directly test how much a network's performance is due to superposition.

\textbf{Improving interpretability}: We could choose an $L_1$ penalty for the norm constraint on the modifier vector (instead of the $L_2$ norm);  or we could constrain $W_1$ to be low rank; alternatively, we could quantize the output of $W_1 x$ in order to put hard limits on the amount of superposition a network can do.

}

\subsection{Feature construction in bilinear layers decompose into a sum of pairwise interactions between input features}\label{sec:pairwise}

Not all layers have the same `amount' of nonlinearity. Some are more nonlinear than others. Here we characterize the amount of nonlinearity layers can have, which sheds light on how bilinear layers differ from standard MLPs.

Let $C(d_i^I , d_j^O, a^I)$ quantify the contribution of input feature $d_i^I \in D^I$ to the activation (or `selection') of output feature $d_j^O \in D^O$. We then have the following (non-comprehensive) set of degrees of nonlinearity.

\begin{itemize}
    \item \textbf{Linear}: Fully linear layers have no nonlinearity. There are therefore no interactions between input features during feature construction (since there is no modifier vector). The amount that input feature $d_i^I$ contributes to the selection of output feature $d_j^O$ is quantified simply as $C(d_i^I , d_j^O, a^I) = [W d_i^I a^I_i]^\top d_j^O$, which is just the inner product between the preactivation caused by that input feature and the output feature.
    \item \textbf{Additively pairwise nonlinear}: In this case, output features are determined by a sum of pairwise interactions between features. For example, if input features $d_1^I, d_2^I, d_3^I$ are active in the input, the contribution of $d_i^I$ (where $i \in \{ 1, 2, 3\}$) to each output feature can be described as a sum of pairwise nonlinear interactions, $C(d_i^I , d_j^O, a^I) = [f(d_i^I; d_1^I, a_1^I) + f(d_i^I; d_2^I, a_2^I) + f(d_i^I; d_3^I, a_3^I)]^\top d_j^O$, where $f(\cdot)$ is some nonlinear function of the two interacting features.
    \item \textbf{Fully nonlinear}: The contribution an input feature makes to the selection of an output feature depends on every other feature in a way that can't be decomposed into a sum. The contribution of $d_i^I$ to each output feature can only be described as an all-to-all nonlinear interaction between input features that cannot be broken down into linear components: $C(d_i^I , d_j^O, a^I) = g(d_i^I; d_1^I, d_2^I, d_3^I, a^I)^\top d_j^O$, where $g(\cdot)$ is some (non-additively-pairwise) nonlinear function.
\end{itemize}

The task of understanding additively pairwise nonlinearity is easier than full nonlinearity because we can study each pairwise interaction between features and sum them up. Understanding full nonlinearity is significantly harder because there is no way to linearly decompose the function $g$. Sadly, standard MLPs are fully nonlinear. However, we show that bilinear layers are additively pairwise nonlinear, making them significantly easier to analyze. 

Suppose the input to a bilinear layer $x'$ consists of a linear combination of two input features $d_1^I$ and $d_2^I$ , i.e. $x'=a_1 d_1^I + a_2 d_2^I$. Using the re-expression of the bilinear layer, inputting $x'$ into equation \ref{bilin_rexpr} yields
\begin{multline}
(a_1 d_1 + a_2 d_2) \boldsymbol{\cdot}_{1 2} B \boldsymbol{\cdot}_{2 1} (a_1 d_1 + a_2 d_2) = \\
a_1 d_1 \boldsymbol{\cdot}_{1 2} B \boldsymbol{\cdot}_{2 1}  a_1 d_1 + \\ a_1 d_1 \boldsymbol{\cdot}_{1 2} B \boldsymbol{\cdot}_{2 1} a_2 d_2 + \\ a_2 d_2  \boldsymbol{\cdot}_{1 2} B \boldsymbol{\cdot}_{2 1} a_1 d_1 + \\ a_2 d_2  \boldsymbol{\cdot}_{1 2} B \boldsymbol{\cdot}_{2 1} a_2 d_2 \quad \\
\end{multline}

More generally, for arbitrary linear combinations of input features:
\begin{multline}\label{eq:pairwise_eq}
(W_1 x) \odot (W_2 x) = (\sum_{i \in R} a_i d_i) \boldsymbol{\cdot}_{1 2} B \boldsymbol{\cdot}_{2 1} (\sum_{i \in R} a_i d_i) = \sum_{i \in R} \sum_{j \in R} a_i a_j d_i \boldsymbol{\cdot}_{1 2} B \boldsymbol{\cdot}_{2 1} d_j
\end{multline}

where $R$ is the set of indices of nonzero feature coefficients. Equation \ref{eq:pairwise_eq} shows that, although all features interact to determine the output features, these interactions can be understood as a sum of pairwise interactions between features. Hence bilinear layers are only additively pairwise nonlinear. 

We hope that this simplicity can be leveraged to tell simple stories about how particular input features (hopefully sparsely) activate particular default output features and modifier features. Then, if we understand the functional properties of those default output features and the kinds of functional modifications that those modifier features make, then we may be able to understand the properties of the output features.

\subsection{How should we study feature construction?} \label{how_to_study_interactions}
At this early stage, it's not totally clear how best to analyze the structure of bilinear networks. What is clear is that doing so will be easier than analyzing fully nonlinear computations, since we're simply studying the structure of tensors, which is a relatively well understood domain in mathematics. In advance of empirical results, I speculate on a few non-mutually exclusive ways to proceed in this section.
\begin{enumerate}
   \item \textbf{Large coefficients of $B$}:
   As discussed at the beginning of section \ref{bilinear_intro}, when interpreting any linear transformation, there may be so many coefficients that it may be necessary to prioritize our analyses by studying only the largest coefficients.
   One way to leverage this is simply to study the largest coefficients of $B$ and how they would influence interactions between commonly observed pairs or groups of input features.
   \item \textbf{Tensor decomposition}:
   Building on (1), we could perform Higher Order Singular Value Decomposition (\href{https:\/\/en.wikipedia.org\/wiki\/Higher-order\_singular\_value\_decomposition}{HOSVD}) and study the structure of the most influential ranks of the tensor.
   \item \textbf{Maximally modified default output features}:
   Recall that one way to view the bilinear network is that one side of the elementwise multiplication modifies the linear transformation on the other side. This suggests an way to prioritize the analysis of how particular features are constructed: For each input feature, we should prioritize analysis of the most modified default output features. 
   Concretely, define $$U^{(2, d_i)} := d_i^\top W_1 \boldsymbol{\cdot}_{1 2} Z \boldsymbol{\cdot}_{2 1} U^{(2)}.$$ This is the set of output features caused by the modifications that input feature $d_i$ makes to default output feature $U^{(2)}$.
   Then, for each input feature $d_i$ we should study the top k most modified default output features, i.e. 
   \begin{equation} 
   \underset{l}{\text{arg top-k}} ( ||U^{(2)}_{:,l} -U^{(2, d_i)}_{:,l} || ) 
   \end{equation}
   This would let us focus on the most significant modifications that a given input feature makes to the default output features.
   But we can prioritize our analyses further than that. The modifications that an input feature makes to the default output features don't matter unless the default output feature is actually activated by that feature or some other feature that is simultaneously present in $x$. Therefore we can identify pairs of features, $(d_l, d_m)$ that are correlated (or that have often appeared at the same time) and where $U^{(2, d_l)}$ is both one of the default output features that is most modified by $d_m$ and simultaneously one of the default output features that is most activated by $d_m$.
\end{enumerate}

\section{Conclusion}\label{Conclusion}

The simplicity of bilinear layers makes formal analysis much easier than for standard MLPs. One of the most important things bilinear layers give us are analysable expressions for performant transformers with both attention heads and MLP layers. I hope that this will eventually let us formally analyze the structure of the representations of large language models in this class. This might reveal interesting features and circuits in a similar way that the mathematical framework for attention-only transformers introduced by \cite{math_framework} helped to reveal reveal QK- and OV-circuits, induction heads, and virtual attention heads. Curiosity aside, an expression for models with bilinear layers may let us make stronger claims about safety. For instance, it may let us more directly compare circuit structure in different models, and enable us to make inferences about model behaviour without necessarily running the model. 

Another potential research direction is analyzing learning dynamics. Models with bilinear layers seem like they might lend themselves to mathematical analysis in a similar fashion to the deep linear layers studied by \cite{saxe2013exact}. Learning dynamics may be important for safety, since understanding them may be necessary to be able to predict dangerous model behaviors before they emerge.

Lastly, and most speculatively, bilinear layers offer the potential to understand the mechanisms of feature construction, which may be necessary for understanding a potentially exponentially large number of features represented in language models. There is still much empirical work to do to evaluate whether intuiting the mechanisms of feature construction is possible. Overall, I hope that this note might pique the interest of the interpretability community by highlighting an architecture that is much gentler on the intuitions than standard MLPs. 

\section*{Acknowledgements}
I thank Trenton Bricken for helpful discussions that initiated my search for layers that could be described in terms of higher order tensors. I thank Beren Millidge, Sid Black, and Dan Braun for helpful discussions and detailed feedback on this work.

\bibliography{refs}

%%%%%%%%%%%%%%%%%%%%%%%%%%%%%%%%%%%%%%%%%%%%%%%%%%%%%%%%%%%%

%%%%%%%%%%%%%%%%%%%%%%%%%%%%%%%%%%%%%%%%%%%%%%%%%%%%%%%%%%%%

\appendix

\section{Tensor inner product examples} \label{sec:inner_prod_examples}

The definition of tensor inner product we use is

$$\boldsymbol{U}^{(n)} \boldsymbol{\cdot}_{jk} \boldsymbol{V}^{(m)} = \boldsymbol{T}^{(n+m-2)}$$ where 

$$
\boldsymbol{T}_{\gamma_1 \cdots \gamma_{j-1} \gamma_{j+1} \cdots \gamma_{n}  \gamma'_{1} \cdots \gamma'_{k-1} \gamma'_{k+1} \cdots \gamma'_{m} } = \sum_\beta \boldsymbol{U}_{\gamma_1 \cdots \gamma_{j-1} \beta \gamma_{j+1} \cdots \gamma_n}  \boldsymbol{V}_{\gamma'_{1} \cdots \gamma'_{k-1} \beta \gamma'_{k+1} \cdots \gamma'_{m}}
$$

\textbf{Example 1}: $\bold{U}^{(1)} \bold{\cdot}_{11} \bold{V}^{(1)} = \bold{T}^{(0)} = \sum_\beta \bold{u}_\beta \bold{v}_\beta = \bold{u}^\top \bold{v}$, which is just the standard inner product, resulting in a scalar.

\textbf{Example 2}: $\bold{U}^{(2)} \bold{\cdot}_{21} \bold{V}^{(1)} = \bold{T}^{(1)} $ where $ \bold{T}_i = \sum_\beta \bold{U}_{i\beta} \bold{v}_\beta$. This is multiplication of a matrix on the right and a vector on the left: $ \bold{T} = \bold{U} \bold{v}$.

\textbf{Example 3}: $\bold{U}^{(2)} \bold{\cdot}_{11} \bold{V}^{(1)} = \bold{T}^{(1)} $ where $ \bold{T}_i = \sum_\beta \bold{U}_{\beta i} \bold{v}_\beta$. This is equivalent to multiplication of a transposed matrix on the left and a vector on the right: $ \bold{T} = \bold{U}^\top \bold{v} $.

\textbf{Example 4}: $\bold{U}^{(1)} \bold{\cdot}_{12} \bold{V}^{(2)} = \bold{T}^{(1)} $ where $ \bold{T}_i = \sum_\beta \bold{u}_{\beta} \bold{V}_{i\beta} $. This equivalent to multiplication of transposed vector on the left and a matrix on the right: $ \bold{T} = \bold{u}^\top \bold{V}^\top $. Note that $\bold{T}$ is a rank one tensor, so $\bold{T} = \bold{u}^\top \bold{V}^\top = \bold{V} \bold{u}$ since tensor notation disposes of the convention that vectors are column vectors or row vectors; instead they are just rank-one tensors. We somewhat abuse notation in this work by assuming standard vector-matrix conventions for multiplication unless the tensors we're dealing with are rank-three or above, in which case we use tensor inner product notation.

\textbf{Example 5}: $\bold{U}^{(3)} \bold{\cdot}_{11} \bold{V}^{(1)} = \bold{T}^{(2)}$ which is the matrix that is a sum of matrices consisting of slices of the rank-three tensor $\bold{T} = \sum_\beta \bold{U}_{\beta::}  \bold{v}_{\beta}$. If we imagine the rank-three tensor as a cube, this example flattens the tensor along its height by taking the inner product between $\bold{v}$ and every 3-d column of $\bold{U}$. 

\textbf{Example 6}: $\bold{U}^{(2)} \bold{\cdot}_{23} \bold{V}^{(3)} = \bold{T}^{(3)}$ which is the rank-three tensor where  $ \bold{T}_{i::} = \sum_\beta \bold{U}_{:\beta} \bold{V}_{i:\beta} $. If we imagine tensor $\bold{V}$ as a cube, here take each front-to-back-row and get its inner product with the corresponding row $i$ of matrix $\bold{U}$.

\end{document}